# FROM PERCEPTION TO CONTROL: AN AUTONOMOUS DRIVING SYSTEM FOR A FORMULA STUDENT DRIVERLESS CAR

[1]Tairan Chen*; [1]Zirui Li; [1]Yiting He; [1]Zewen Xu; [1]Zhe Yan; [1]Huiqian Li
[1] Beijing Institute of Technology, China

ABSTRACT
This paper introduces the autonomous system of the "Smart Shark II" which won the Formula Student Autonomous China (FSAC) Competition in 2018. In this competition, an autonomous racecar is required to complete autonomously two laps of unknown track. In this paper, the author presents the self-driving software structure of this racecar which ensure high vehicle speed and safety. The key components ensure a stable driving of the racecar, LiDAR-based and Vision-based cone detection provide a redundant perception; the EKF-based localization offers high accuracy and high frequency state estimation; perception results are accumulated in time and space by occupancy grid map. After getting the trajectory, a model predictive control algorithm is used to optimize in both longitudinal and lateral control of the racecar. Finally, the performance of an experiment based on real-world data is shown.

## I. INTRODUCTION

With the development of autonomous driving, the architecture of the automatic driving system has become very important. A good architecture can improve the safety and handling maneuverability of autonomous vehicles. In general, most hardware architectures of autonomous driving are similar to each other, but the intelligence level of self-driving cars varies widely. The key reason is the software architecture of autonomous driving is different. Software architecture plays a vital role in the implementation of autonomous driving.

Several autonomous driving system architectures based on autonomous passenger vehicles are proposed. Autoware is the world's first "all-in-one" open-source software for self-driving vehicles [1]. The capabilities of Autoware are primarily well-suited for urban cities; Apollo is a high-performance, flexible architecture of autonomous vehicles. One of the Apollo project's goals is to create a centralized place for original equipment manufacturers (OEMs), startups, suppliers, and research organizations to share and integrate their data and resources [2]. In addition, the autonomous architectures in Google, Tesla and Nvidia are also developed for autonomous passenger vehicles which are also not relevant to the FSD competition.

Recently, as autonomous ground vehicles (AGVs) attracted focus in both academic and commercial applications, combining AGV with Formula Student Competition (FSC) is significant for education purpose and the development of Formula Student Competition. Some self-driving racecars are developed by students. Software algorithms are applied to achieve high speed driving on a previously unknown racetrack. The first autonomous racecar to win a Formula Student Driverless competition is [3]. TUW Racing [4] covers the framework, the sensor setup and the approaches.

The "Smart Shark II" autonomous system uses a 3D LiDAR and a monocular camera to perceive its surroundings. In addition, a GPS/INS and wheel speed sensors are added for state estimation. All the data are processed by an industrial computer with GTX 1050Ti running ROS and a rapid prototype ECU which can be programmed in Simulink environment.

The rest of this paper is organized as follows: Section II introduces the autonomous system architecture of Smart Shark II in detail. Section III presents and discusses the results of experiments. Conclusion and future work are presented in section IV.

II. ARCHITECTURE OF AUTONOMOUS SYSTEM

In this section, we present a detailed description of the architecture of the autonomous system of the Smart Shark II, including EKF-based localization, robust perception, occupancy grid map and Model Predictive Control (MPC). The system structure is shown in Fig. 1:

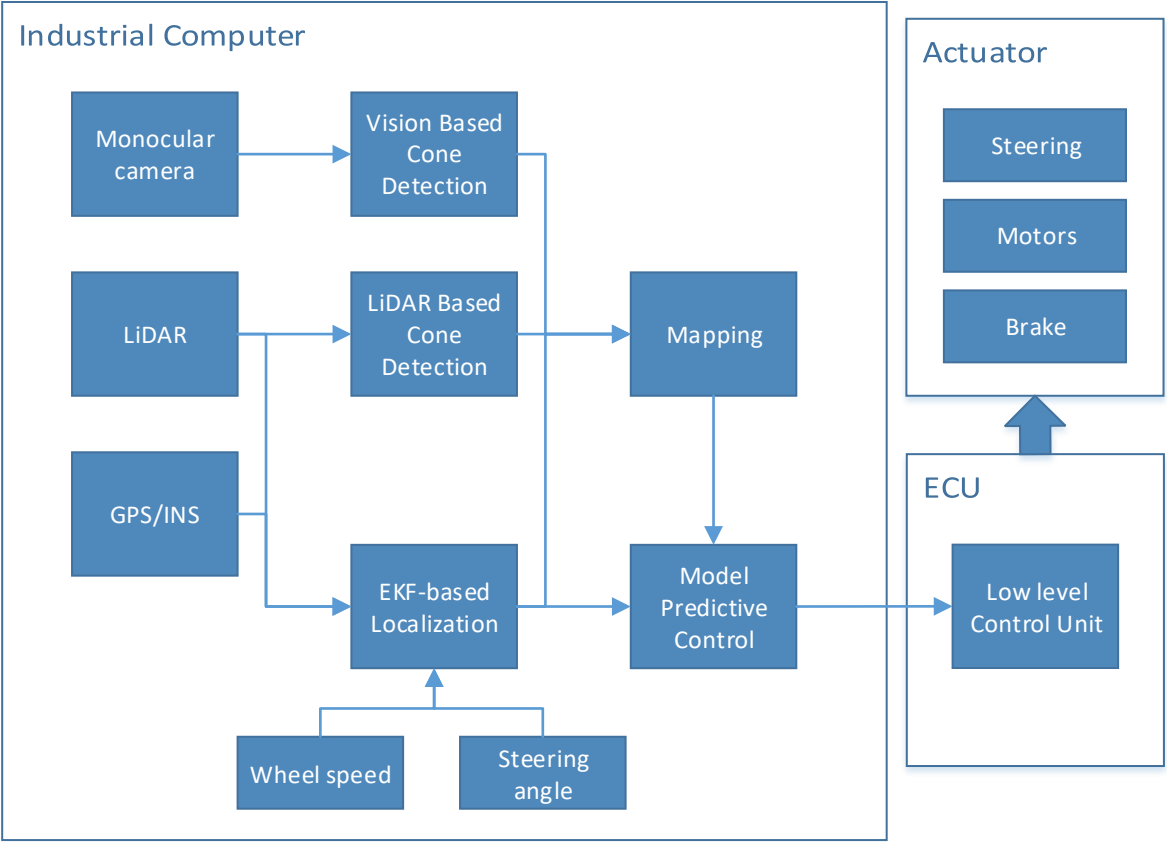

Fig. 1 Autonomous system structure diagram

## A. EKF-based localization

In the "Smart Shark II" localization system, GNSS provides global position (latitude, longitude and altitude) of the racecar with high accuracy but low update frequency; INS can offer speed and angle acceleration with rough accuracy but high update frequency; inspired by LOAM [5], a real-time method for LiDAR odometry can also provide the accurate position. Multiple sensors can be fused to provide more accurate and higher frequency localization. With the Extended Kalman Filter (EKF) algorithm, redundant but highly uncertain measurement information can be combined into a more reliable information with less uncertainty. In addition, it can be more reliable in the event of sensor failure. As long as more than one sensor is reliable, the localization module can continue working well. The major derivation of the EKF is described below and its architecture is shown in Fig. 2:

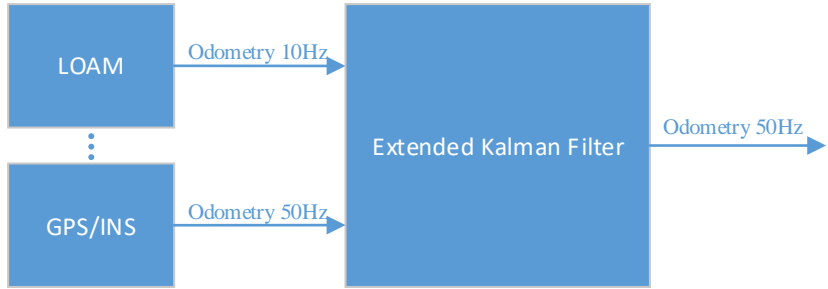

Fig. 2: EKF-based localization architecture

Our goal is to estimate the 2D pose and velocity of the racecar with multiple sensors. The process can be described as a nonlinear dynamic system as follows:

$$\mathbf{x_k} = f(\mathbf{x_{k-1}}) + \mathbf{n_{k-1}} \tag{1}$$

Where $x_k$ represents the state of the racecar at time k. $f$ is a nonlinear state transfer function and $\mathbf{n_{k-1}}$ is the process noise that subjects to normal distribution. The state vector $x_k \in \mathbb{R}^{6\times1}$ is defined as

$$\mathbf{x_k} = [\mathbf{p}^T, \theta, \mathbf{v}^T, r]^T \tag{2}$$

where $\mathbf{p} = [x, y]^T$ and $\theta$ represent the position and heading of the car, $\mathbf{v} = [v_x, v_y]^T$ and $r$ represent the linear and angular velocities respectively. The nonlinear state transfer function is defined as:

$$\begin{aligned} \dot{\mathbf{p}} &= \mathbf{R}(\theta)^T \mathbf{v} \\ \dot{\theta} &= r \\ \dot{\mathbf{v}} &= \mathbf{a} + [v_y r, -v_x r]^T + \mathbf{n_v} \\ \dot{r} &= n_r \end{aligned} \tag{3}$$

where $\mathbf{a}$ is the linear acceleration, $\mathbf{R}(\theta)$ is the 2D rotation matrix between the vehicle body frame and the world reference frame. $\mathbf{n_v}$ and $n_r$ is the process noise that subjects to normal distribution.

In order to update the state with the measured values, we will take the measured values obtained by the sensor as follows:

$$\mathbf{z_k} = h(\mathbf{x_k}) + \mathbf{v_k} \tag{4}$$

Where $\mathbf{z_k}$ represents the measured value of the sensor at time $k$, $h$ is a nonlinear sensor model that maps the state into measurement space, and $\mathbf{v_k}$ is the measurement noise that subjects to the normal distribution.

As described above, the step of the EKF is to recursively predict and correct the state of the racecar. The first step of the algorithm, as shown in equations (5) and (6), is a prediction step to predict the current state and error covariance:

$$\hat{\mathbf{x}} = f(\mathbf{x_{k-1}}) \tag{5}$$

$$\widehat{P_k} = FP_{k-1}F^T + Q \tag{6}$$

In the above expression, *f* is a three-dimensional dynamic model derived from Newtonian mechanics, which is used to estimate the state, the estimated variance $\widehat{P_k}$ is the covariance $Q$ of the random noise superimposed on the Jacobian matrix $F$ of *f*.

The equations (7) through (9) is used to correct the estimated state and covariance:

$$K = P_k H^T (H P_k H^T + R)^{-1} \tag{7}$$

$$x_k = \widehat{x_k} + K(z - H\widehat{x_k}) \tag{8}$$

$$P_k = (1 - KH)\widehat{P_k}(1 - KH)^T + KRK^T \tag{9}$$

The observation matrix H, measurement covariance R and $\widehat{P_k}$ are used to calculate the Kalman gain K; then use the Kalman gain to update the state vector and covariance matrix; the Joseph form covariance is used to update the formula (8) to ensure that $P_k$ is positive semi-definite.

B. LiDAR-based Cone Detection

A LiDAR-only method in perception is proposed based on the rules of the FSAC, in which the color estimation problem is transferred into a geometric distribution problem. This method means that if the cone placement does not meet the rules, color estimation errors will occur, but it still has value for analyzing the distribution of road boundaries. First an obstacle detection method is described. After knowing the obstacle position, a convolutional neural network (CNN) is designed to detect the distribution of blue and red cones. The LiDAR-based Cone Detection architecture is depicted in Fig. 3, the key methods of which are described below.

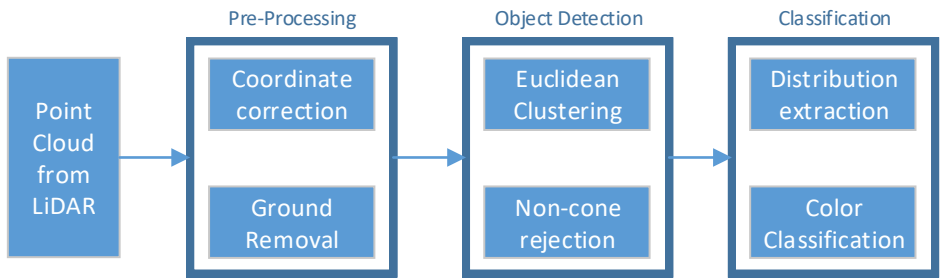

Fig. 3 Flow diagram for cone position and color estimation based on geometric distribution

1) LiDAR-based obstacle detection

Before objects detection, the ground points from the raw point cloud should be removed. We use an adaptive ground removal algorithm [6], which adapts to changes in inclination of the ground. In this algorithm, ground is split into multiple sectors and bins; lines are fit through the lowest points of each bin. Finally, all points within a threshold of the closest line are removed. The point cloud after filtering out the ground is shown in Fig. 4 (a).

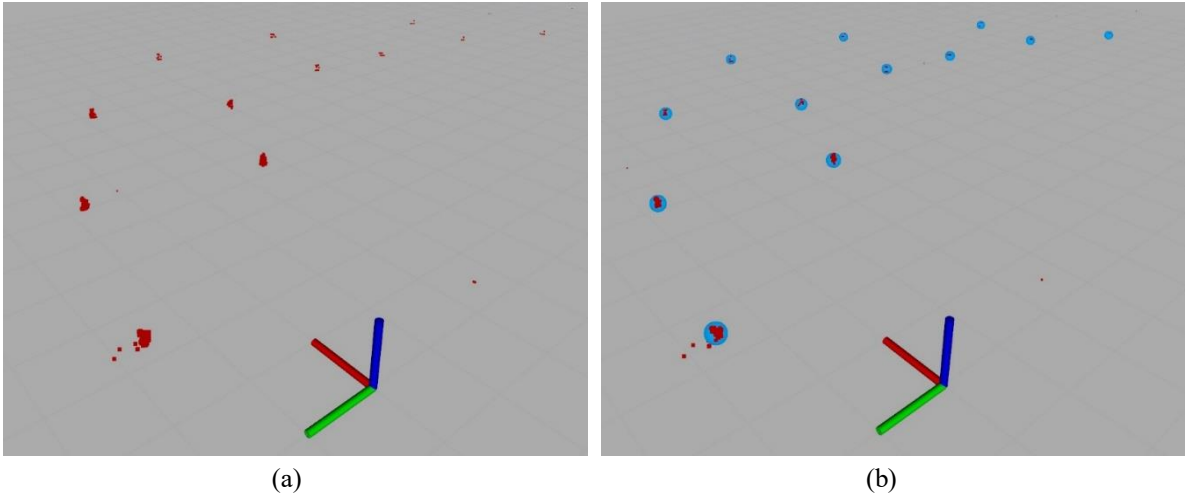

(a)                                              (b)

Fig. 4: (a) Points cloud after removing the ground. (b) The blue spheres show the cones cluster result.

Once the ground point cloud is removed, the centroid position of a set of point clouds is obtained by a clustering algorithm. By calculating the envelope size of the clustered object, it is possible to filter out obstacles different in size from the cone. Finally, the obstacle position similar to the cone size is obtained as shown in Fig. 4 (b).

2) CNN cones' distribution extraction

After the cluster and filtration by size, the position of cones can be detected as shown in Fig. 6 (a). However, the color of cones is unknown. To estimate the color of cones, a neural network is developed. Inspired by the different color of cones on the left and right boundary of the track, this neural network is used to distinguish the geometric distribution of cones. The architecture of the convolutional neural network consists of four convolution layers, each layer followed by a rectified linear unit (ReLU) as the non-linear activation. Finally, there is a fully-connected layer that classifies the color of the cones. The network structure is shown in Fig. 5.

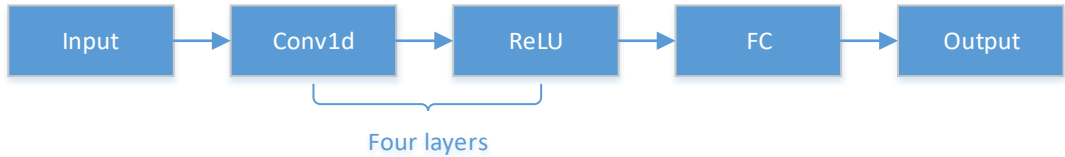

Fig. 5 CNN structure of cones' distribution extraction

For data standardization, the position of cones coordinates (x, y) is added into a 30-dimensional vector as the input. If the number of cones is less than 15, coordinate (0, 0) will be filled into the vacant position of vector to satisfy the vector like:

$$[x_1, y_1, x_2, y_2, \ldots, x_i, y_i, \ldots, x_{15}, y_{15}]$$

Each input coordinate corresponds to a classification label, red as 1, blue as 2, and (0,0) as 0. The classification result of cones is shown in Fig. 6 (b).

Cone positions and color labels were obtained from a simulation environment. The data were split as 6000 items for training, 2,000 for validation and 2000 for testing. Adaptive Moment Estimation (Adam) was used for optimization, with a learning rate, lr = 0.001 and a batch size of 16. The network is trained for 200 epochs. The network is implemented in PyTorch and applied to ROS via libtorch.

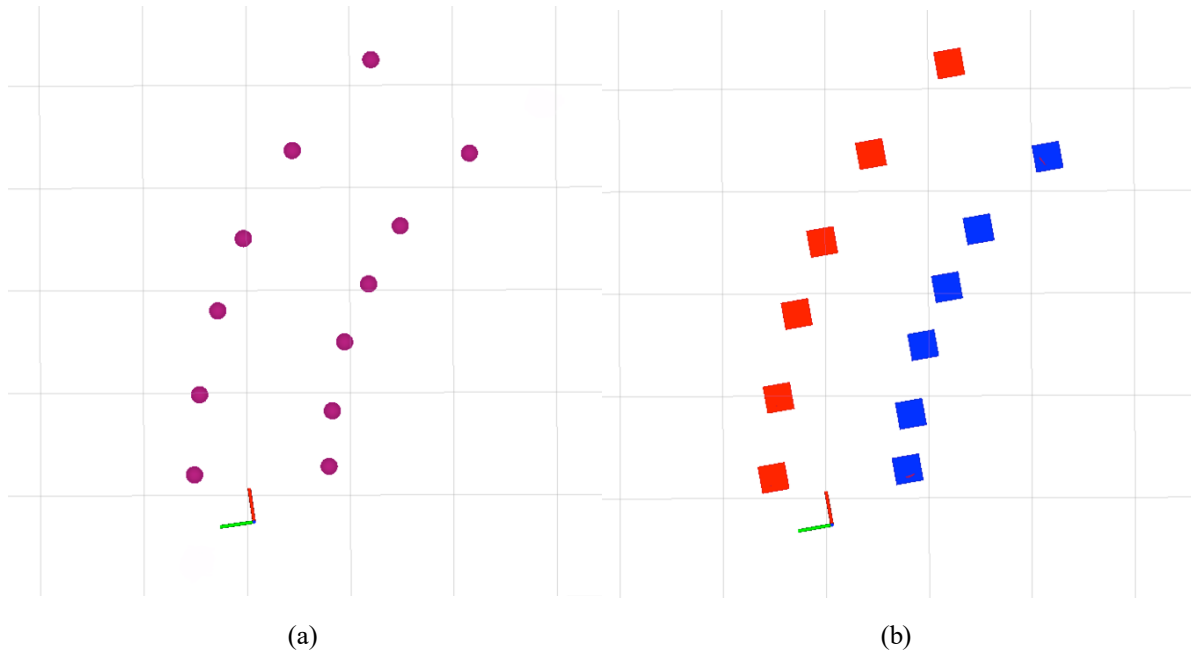

(a)　　　　　　　　　　　　　　　　　　(b)

Fig. 6: (a) the points cloud clustering results without color is represented by purple spheres, (b) the color of cones predicted by neural network are represented by red and blue squares.

C. Vision-based Cone Detection and Pose Estimation

A monocular camera is used to detect the color of cones and to estimate the position of cones by joint calibration with the LiDAR. Using YOLOv3 [7] as the target detection algorithm, the cones' color and the positions in the image are detected by training the blue and red cones with bounding boxes. Moreover, through the special boundary frame labeling method, the midpoint position of the bottom edge of the cone is conveniently obtained, which is beneficial to directly estimate the position of cones.

1) Cone Detection

YOLOv3 is used in cone detection, which offers good accuracy and performance as shown in Fig. 7 (a). It uses an entire image as input to the network, directly returning to the output layer and the class to which it belongs. The red and blue cones in the image are used for training. In order to get the midpoint of the bottom edge of the cone, we select this special point as the bottom midpoint of the bounding box when labeling the dataset. When estimating the position of cones, the midpoint of the bottom edge of the bounding box can be directly selected as the contact point between the cone and the ground which is shown in Fig.7 (b). Different ground and lighting conditions are chosen to improve the robustness of cone detection.

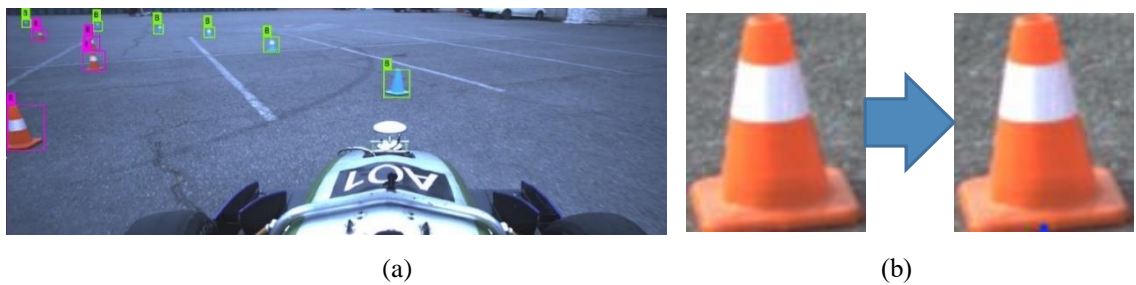

(a)　　　　　　　　　　　　　　　　　　(b)

Fig. 7: (a) Bounding boxes from YOLOv3 in image from monocular camera, (b) The point where the cone is in contact with the ground is marked by a blue dot, which happens to be the midpoint of the bottom edge of the picture.

2) Pose Estimation

We assume that the ground is flat. To estimate the position of the cone, a joint calibration of LiDAR and camera algorithm [8] is applied. The perspective matrix can be constructed between the two planes by choosing four pairs of different corresponding points:

$$[x', y', w'] = [u, v, w] \begin{bmatrix} a_{11} & a_{12} & a_{13} \\ a_{21} & a_{22} & a_{23} \\ a_{31} & a_{32} & a_{33} \end{bmatrix} \quad (10)$$

$$x = \frac{x'}{w'}, y = \frac{y'}{w'} \quad (11)$$

Where: $u$ and $v$ represent the pixel coordinate of the bottom center of the cones. $\frac{x'}{w'}$ and $\frac{y'}{w'}$ represent the position of the cones in XOY plane. $\begin{bmatrix} a_{11} & a_{12} \\ a_{21} & a_{22} \end{bmatrix}$ is the linear transformation. $[a_{13} \quad a_{23}]^T$ is perspective effect segment and $[a_{31} \quad a_{32}]$ is translation vector. Fig. 8 (a) shows four points in the picture we choose to calibrate, Fig. 8 (b) shows the aerial view after transformation.

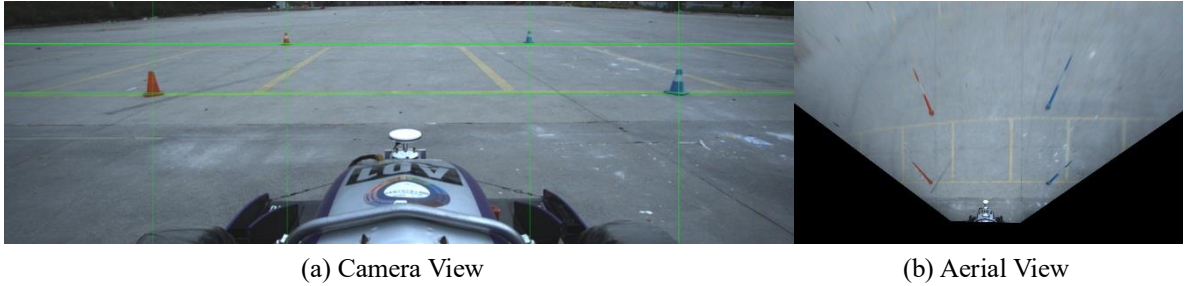

(a) Camera View  (b) Aerial View

Fig. 8. Camera View (a) and Joint calibration by perspective transformation (b).

D. Occupancy Grid Map

Due to the limitation of the angle of the visual sensor image, it is not possible to detect all the cones in front of the racecar at the turning position. Therefore, by constructing a map, the colors and positions of cones are accumulated in time and space to output stable track information in real time.

The construction of occupancy grid map can be combined with LiDAR-based and Vision-based cone detection. In addition, the output of YOLOv3 also contains the probability of the cone color. Through the joint calibration of the LiDAR and the camera, the transformation between the two coordinate systems is obtained, and the coordinate of the visual cone can be transformed into the LiDAR coordinate system.

The map is stored in the form of a grid map with a resolution of 0.1 m. For each grid, there are two states: occupied or blank. $p(s = 1)$ is the probability of a blank state. $p(s = 0)$ is the probability of an occupied state. $Odd(s) = \frac{p(s=1)}{p(s=0)}$ is the state of the grid. When the perception gets a new cone (z) at the next moment, the

method of updating the grid state is as follows:

$$Odd(s|z) = \frac{p(s=1|z)}{p(s=0|z)} \tag{12}$$

According to the Bayes theorem:

$$p(s=1|z) = \frac{p(z|s=1)p(s=1)}{p(z)} \tag{13}$$

$$p(s=0|z) = \frac{p(z|s=0)p(s=0)}{p(z)} \tag{14}$$

The next state is:

$$Odd(s|z) = \frac{p(z|s=1)}{p(z|s=0)} Odd(s) \tag{15}$$

Logarithm of the equation:

$$\log(Odd(s|z)) = \log\left(\frac{p(z|s=1)}{p(z|s=0)}\right) + \log(Odd(s)) \tag{16}$$

Define $\log\left(\frac{p(z|s=1)}{p(z|s=0)}\right)$ as $lomeas$, $\log(Odd(s|z))$ as $S_t$, $\log(Odd(s))$ as $S_{t-1}$.

The status update formula is as follows:

$$S_t = S_{t-1} + lomeas \tag{17}$$

The occupancy grid map created in the global coordinate system is shown in Fig. 9:

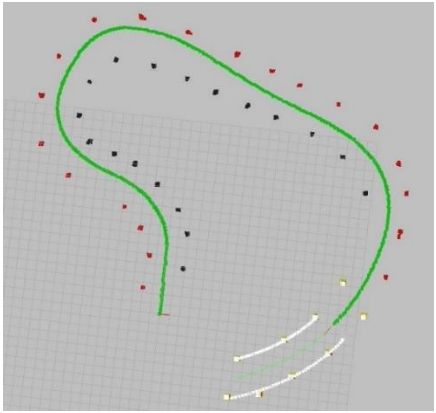

Fig. 9 an occupancy grid map, red pixels represents red cones, black pixels represents blue cones, others represents no cones.

E. <u>Model Predictive Control</u>

After the successful completion of the first lap, the position of all the cones in the entire track will be converted to the longitude and latitude. Therefore, the vehicle can use the integrated navigation to know its position and posture on the track. The task is to let the vehicle track the known path which is the middle line of the track.

In this part we introduce an optimization-based control strategy for path following, which includes dynamic modelling, constraints and the real-time optimization solver.

1) Vehicle dynamics modeling for formula racecar

Shown in Fig. 10, a 3-DOF vehicle model is used in the MPC-based controller as the fundamental model in predicting the longitudinal and lateral control output. The three degrees of freedom are vehicle lateral speed, longitudinal speed, and yaw rate.

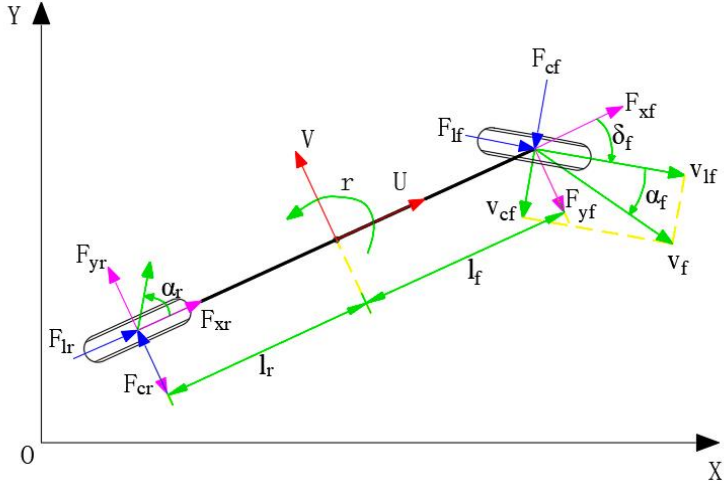

Fig. 10. The 3-DoF vehicle model

Based on the 3-DOF vehicle model, the state-space equation is presented as:
$$\dot{\xi} = f(\xi) + Au_1 + Bu_2 \tag{18}$$

With $\xi = \begin{bmatrix} x \\ y \\ \psi \\ U \\ V \\ r \\ \delta_f \\ a_x \\ e_y \\ e_\psi \end{bmatrix}$, $f(\xi) = \begin{bmatrix} U\cos\psi - V\sin\psi \\ U\sin\psi - V\cos\psi \\ r \\ a_x \\ (F_{yf} + F_{yr})/M - Ur \\ (F_{yf}l_f - F_{yr}l_r)/I_{zz} \\ 0 \\ 0 \\ Ue_\psi + V \\ r \end{bmatrix}$, $A = \begin{bmatrix} 0 & 0 \\ 0 & 0 \\ 0 & 0 \\ 0 & 0 \\ 0 & 0 \\ 0 & 0 \\ 1 & 0 \\ 0 & 1 \\ 0 & 0 \\ 0 & 0 \end{bmatrix}$, $B = \begin{bmatrix} 0 & 0 \\ 0 & 0 \\ 0 & 0 \\ 0 & 0 \\ 0 & 0 \\ 0 & 0 \\ 0 & 0 \\ 0 & 0 \\ 0 & 0 \\ 0 & -U \end{bmatrix}$, $u_1 = \begin{bmatrix} \varsigma_f \\ J_x^T \end{bmatrix}$, $u_2 = \begin{bmatrix} 0 \\ \kappa \end{bmatrix}$.

In the above expression, $u_1$ is control input and $u_2$ is auxiliary input. The physical meaning of each parameter is shown in Table 1.

TABLE I. Each parameter's physical meaning

| Parameters | Value |
|---|---|
| $(x, y)$ | Center of gravity in global coordinates |
| $\varphi$ | Vehicle's heading angle |
| $U$ | Longitudinal speed |
| $V$ | Lateral speed |
| $r$ | Yaw rate |
| $\delta_f$ | Steering angle |
| $a$ | Longitudinal acceleration |
| $\varsigma_f$ | Steering rate |
| $J_x$ | Longitudinal jerk |
| $M$ | Vehicle's mass |
| $I_{zz}$ | Moment of inertia |
| $l_f$ | Distances between the vehicle's center of gravity and the front axle |
| $l_r$ | Distances between the vehicle's center of gravity and the rear axle |
| $F_{yf}$ | Tire lateral forces generated at the front axle |
| $F_{yr}$ | Tire lateral forces generated at the rear axle |
| $e_y$ | Lateral path following error |
| $e_\psi$ | Heading deviation |

The front and rear tires' slip angles can be calculated by:

$$\alpha_f = tan^{-1}\left(\frac{V + l_f r}{U}\right) - \delta_f \tag{19}$$

$$\alpha_r = tan^{-1}\left(\frac{V - l_r r}{U}\right) \tag{20}$$

The differential equation of vehicle dynamics is a constraint on the dynamics of autonomous formula racecar. In addition, in the process of solving the dynamical differential equation, some variables with actual physical significance cannot exceed the specific constraints.

Hence, the constraints on steering angle, longitudinal speed and acceleration are defined as following:

$$\delta_{f,min} \leq \delta_f \leq \delta_{f,max} \tag{21}$$
$$U_{min} \leq U(t) \leq U_{max} \tag{22}$$
$$a_{min} \leq U(t) \leq a_{max} \tag{23}$$

Where $\delta_{f,min}$ and $\delta_{f,max}$ are steering angle's constraints, $U_{min}$ and $U_{max}$ are longitudinal speed's constraints, $a_{min}$ and $a_{max}$ are acceleration's constraints.

In addition, the control input of the model described in this paper is the steering angle rate and longitudinal jerk. The constraint on both input is defined as following:

$$\varsigma_{f,min} \leq \varsigma_f \leq \varsigma_{f,max} \tag{24}$$
$$J_{x,min} \leq J_x \leq J_{x,max} \tag{25}$$

Where $\varsigma_{f,min}$ and $\varsigma_{f,max}$ are the steering angle rate's constraints, $J_{x,min}$ and $J_{x,max}$ are defined as longitudinal jerk's constraints.

2) Sideslip constraint

Generally speaking, the lateral side slip of the vehicle first occurs on the rear tire. Therefore, the forced constraint is added to the side Angle of the rear tire, and the constraint of the rear wheel in the X linear tire model is as follows:

$$\left|\frac{V - l_r r}{U}\right| \leq \alpha_{r,lim} \tag{26}$$

Where $\alpha_{r,lim}$ is the constraint of sideslip angle. Shown in Fig. 11, $\alpha_{r,lim}$ let the relation between sideslip angle and lateral force constraint in a linear range.

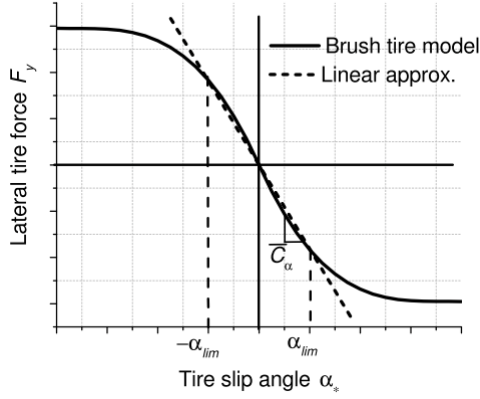

Fig. 11. Brush tire model

3) The environmental constraint

In order to ensure safety, autonomous formula racecar need to be constrained in an appropriate range, which is presented by environment information. Hence, considering the relative position between vehicle and path, the following constraint presents the lateral feasible region, which is defined by maximal and mini- mum lateral path following error.

$$H_{env}\xi^{(k)} \leq G_{env}^{(k)} \tag{27}$$

With $H = \begin{bmatrix} 0 & 0 \\ 0 & 0 \\ 0 & 0 \\ 0 & 0 \\ 0 & 0 \\ 0 & 0 \\ 0 & 0 \\ 0 & 0 \\ -1 & 1 \\ 0 & 0 \end{bmatrix}$, $G_{env}^{(k)} = \begin{bmatrix} e_{y,max}^k - d_s \\ -e_{y,min}^k + d_s \end{bmatrix}$

Where $\xi^{(k)}$ is vehicle's state at step k, $d_s$ is pre-defined comfort distance. The above constraint provides a convex optimization solution for the vehicle to meet the environmental constraints.

4) Cost function

The target of the MPC-based path following controller is to guarantee the vehicle's stabilization and safety under the pre-defined constraints. Therefore, the MPC-based controller can be realized by the cost function:

$$\begin{aligned} J = & \sum_k W_u \left( \delta_f^{(k)} - \delta_f^{(k-1)} \right)^2 \\ & + \sum_k \left( W_{e\psi}(e_\psi^{(k)})^2 + W_{ey}(e_y^{(k)})^2 \right) \\ & + \sum_k W_{sh} S_{sh}^{(k)} W_{sh} \end{aligned} \tag{28}$$

Where the first part establishes the control stability, the second part enforces penalty on the vehicle states deviation, and the third part is slack variables violations.

5) Solve the OCP

The hp-adaptive pseudospectral method is applied for solving the nonlinear OCP of the trajectory planning program. The OCP problem is transcribed into nonlinear programming problem (NLP) by approximating the state and control input. GPOPS is an open source software package that uses the hp-adaptive pseudospectral method to solve OCP. In order to solve the resulted NLP, GPOPS works with the interior point optimizer (IPOPT).

III. EXPERIMENT

This section presents the experiments to test the key components proposed above and analyze the experimental results. Fig.9 shows the location for autonomous driving experiments. It's an open ground and the circular race track for experiments is defined as a loop with 4m width and 5m distance between one pair of cones as Fig. 12 shows.

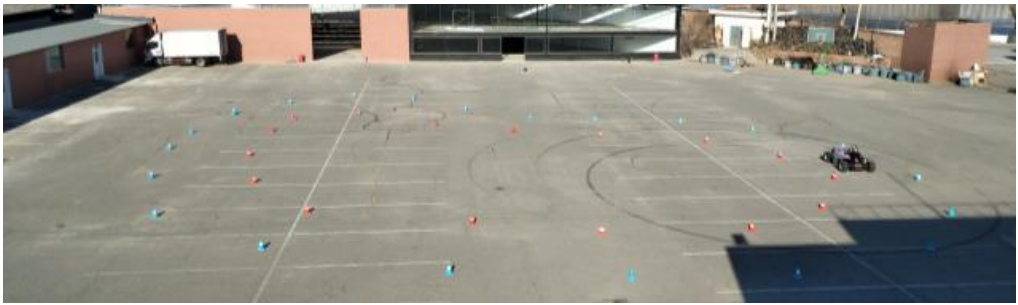

Fig. 12. The test field environment and track for autonomous formula racecar.

A. <u>Robustness of EKF-based localization</u>

Two laps of rounding track are used to test the performance of EKF-based localization. LiDAR Odometry and GPS Odometry are fused by EKF algorithm as shown in Fig. 13 (a). During driving, a ROS node of LiDAR is designed to automatically shutdown to simulate the robustness of localization in the event of sensor failure as shown in Fig. 13 (b). After the LiDAR fails, the system can still provide stable localization.

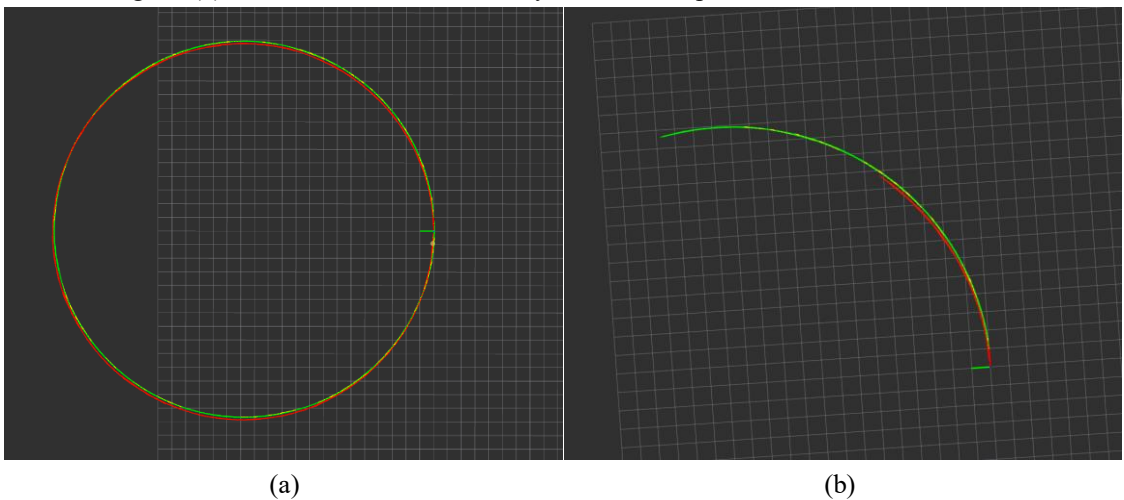

(a)                                                                                       (b)

Fig. 13: (a) two laps of circular track, at each moment, the red arrow represents the LiDAR odometry, the green

arrow represents GPS odometry, the yellow arrow represents fusion odometry by EKF. (b) Fusion odometry after LiDAR failure.

B. <u>Cone Detection</u>

In the first lap, the racecar will operate using LIDAR-based and Vision-based cone detection simultaneously as perception without any prior data of the track. The CNN approach applied to LiDAR is compared with Vision-based cone detection through the real-world datasets which are not used for training as shown in Fig. 14.

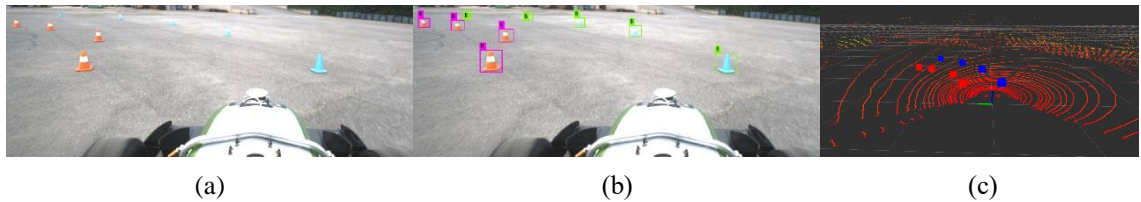

(a)  (b)  (c)

Fig. 14: (a) Original image from monocular camera. (b) Image with YOLOv3 bounding box. (c) Cone point cloud after CNN classification.

Furthermore, the LiDAR color estimation is capped at 10 m. Fig. 15 compares the performance of the CNN classification and YOLOv3. The correct rate of CNN method in the range of 10 meters is 76.7573%.

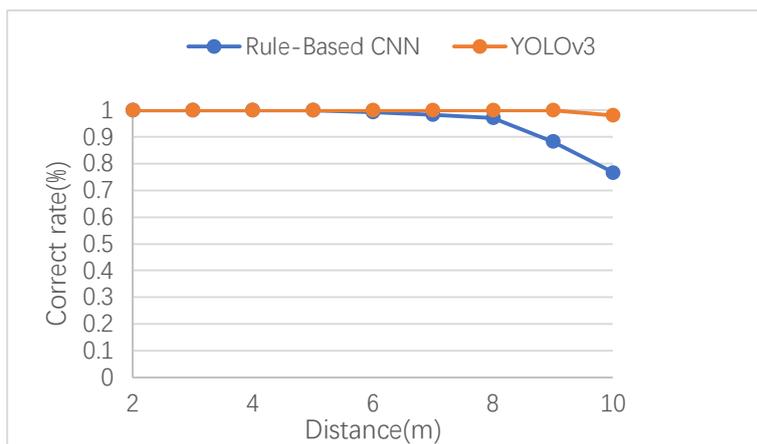

Fig. 15 Classification performance of the CNN and YOLOv3 in the real-world datasets. The accuracy of both methods is sufficient to provide a stable perception.
This figure shows that the CNN method will reverse the classification of red and blue cones at 23.2% in the range of 10m, which is caused by the complex geometric distribution of the cones at the turn. This shortcoming can be overcome by constructing an occupancy grid map and accumulating the perceived results in time and space.

C. <u>Model Predictive Control</u>

The authors' team won the FSAC champion in China. In the competition, the path tracking controller is designed basing on pure pursuit algorithm. The MPC-based controller applied on the formula racecar proposed in this paper is compared to the pure pursuit algorithm. Steering angle, lateral acceleration, sideslip angle and lateral error during the path tracking behaviors are compared and analyzed.

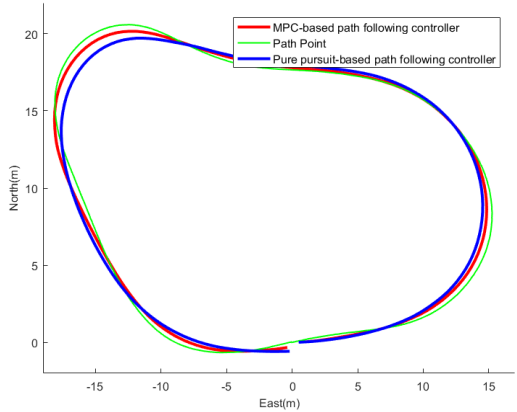

Fig. 16. The trajectory of path following

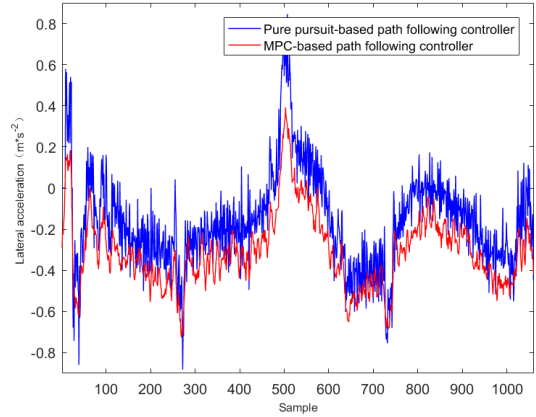

Fig. 17. Lateral acceleration

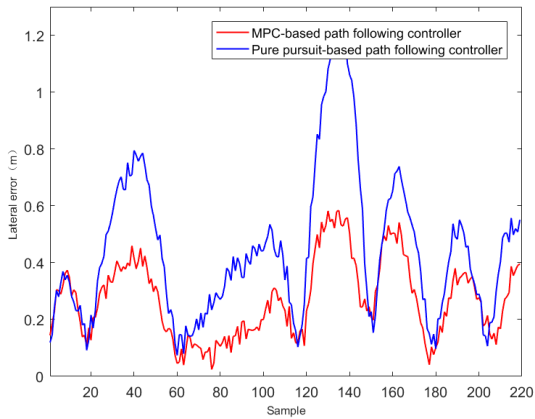

Fig. 18. Lateral error

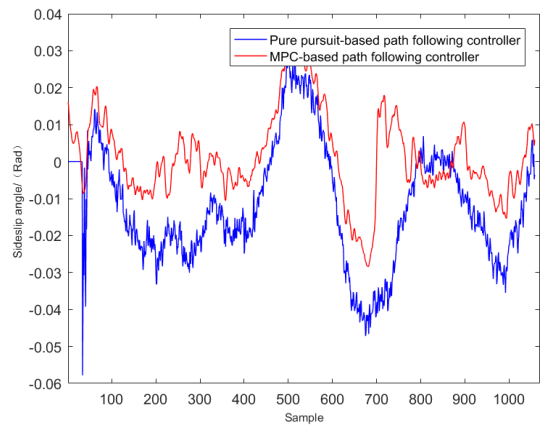

Fig. 19. Sideslip angle

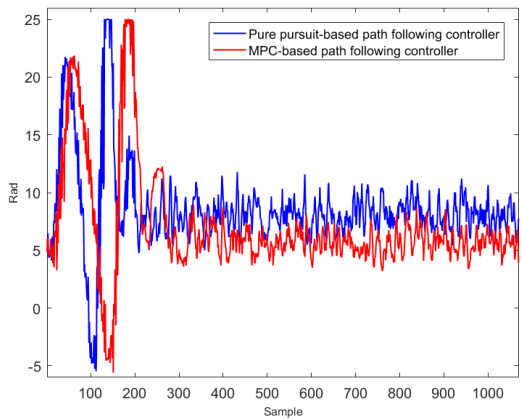

Fig. 20. Steering angle

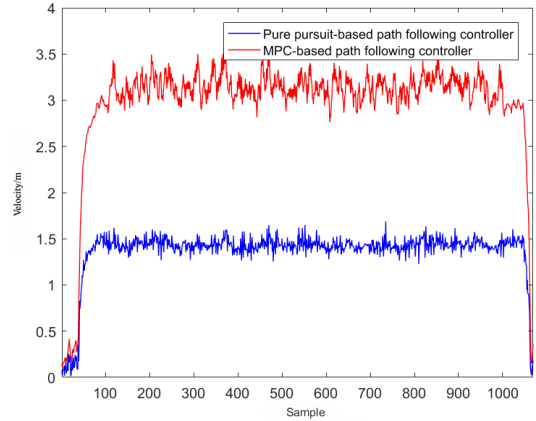

Fig. 21. Velocity

As shown in Fig.16-21, the proposed autonomous formula racecar's path following controller has the same trend with pure pursuit based path following controller in steering angle, lateral acceleration and lateral error. While the proposed controller can keep a higher speed than pure pursuit based controller, which is beneficial to get a high score in the competition. The sample 140 of the pure pursuit algorithm in Figure 18 shows an unusually large error, which we believe is due to poor controller response and inaccurate positioning. For most of the time, the proposed controller has a lower lateral path following error and a smoother lateral acceleration. Table 2 shows the compared results.

TABLE II. compared result of two controller

| Parameters | MPC | Pure pursuit |
|---|---|---|
| Standard deviation of lateral acceleration (m/s$^2$) | 0.1759 | 0.2338 |
| Lateral error of path following(m) | 0.2714 | 0.4520 |
| Average speed(m/s) | 2.9720 | 1.3677 |
| Average sideslip angle(Rad) | 0.0018 | 0.0120 |

IV. CONCLUSION

This paper reports our work on the autonomous system of the "Smart Shark II". For redundant perception, two independent detection methods are applied to the perception system: (a) In a LiDAR-based approach, first we calculate the position of cones by clustering. After that, a CNN-based LiDAR cone detection method solves the color estimation problem by transferring it into a geometric distribution problem. (b) In a Vision-based approach, YOLOv3 detects the cones and the corresponding color. As real-time perception cannot meet high-speed driving, an occupancy grid map accumulates the perceived results in time and space to increase the range and accuracy of perception. To solve the path tracking problem under the rule of FSAC, this paper presents MPC controller on autonomous formula racecar for path following problem in FSAC, which improves the performance compared to the pure pursuit algorithm and provides the possibility to simultaneously travel at its limits. The experiments show the performance of each module, which can race on unknown race tracks at competitive speeds, even if any sensor fails. This architecture of the autonomy system has reference value for other autonomous racecars.

REFERENCES

[1] S. Kato, E. Takeuchi, Y. Ishiguro, Y. Ninomiya, K. Takeda, and T. Hamada. "An Open Approach to Autonomous Vehicles". IEEE Micro, Vol. 35, No. 6, pp. 60-69, 2015.
[2] Claudine Badue, Rânik Guidolini, Raphael V. Carneiro. "Self-Driving Cars: A Survey". 2019
[3] Miguel I. Valls, Hubertus F.C. Hendrikx, Victor J.F. Reijgwart. "Design of an Autonomous Racecar: Perception, State Estimation and System Integration". 2018 IEEE International Conference on Robotics and Automation. May 21-25, 2018
[4] Marcel Zeilinger 1, Raphael Hauk 1, Markus Bader. "Design of an Autonomous Race Car for the Formula Student Driverless (FSD)". 2017
[5] Zhang, J., and S. Singh. "LOAM : LiDAR Odometry and Mapping in real-time." 2014.
[6] M. Himmelsbach, F. v. Hundelshausen, and H. -. Wuensche. "Fast segmentation of 3D point clouds for ground vehicles". In: 2010 IEEE Intelligent Vehicles Symposium. June 2010, pp. 560–565.
[7] Joseph Redmon, Ali Farhadi. "YOLOv3: An Incremental Improvement". 2018
[8] hanqing TIAN, Jun NI, Jibin HU. "Autonomous Driving System Design for Formula Student Driverless Racecar". 2018 IEEE Intelligent Vehicles Symposium. June 26-30, 2018